\documentclass[10pt,twocolumn,letterpaper]{IEEEtran}
\usepackage{cvpr}
\usepackage{times}
\usepackage{epsfig}
\usepackage{graphicx}
\graphicspath{ {./figures/} }
\usepackage{amsmath}
\usepackage{booktabs} 
\usepackage{pdflscape} 
\usepackage{graphicx} 
\usepackage{amssymb}
\usepackage{fancyhdr}
\usepackage{cite}

\fancypagestyle{empty}{%
  \fancyhf{} 
  \cfoot{\thepage} 
}

\usepackage[pagebackref=true,breaklinks=true,colorlinks,bookmarks=false]{hyperref}

\cvprfinalcopy 

\def\cvprPaperID{****} 

\ifcvprfinal\fi
\begin{document}

\title{Utilizing Model Residuals to Identify Rental Properties of Interest: The Price Anomaly Score (PAS) and Its Application to Real-time Data in Manhattan}

\author{Youssef Sultan\\
{\tt\small ysultan@gatech.edu} \\
\and
Jackson C. Rafter\\
{\tt\small jrafter3@gatech.edu} \\
\and
Huyen T. Nguyen\\
{\tt\small hnguyen439@gatech.edu} \\
\and
}
\maketitle

\begin{abstract}
   Understanding whether a property is priced fairly hinders buyers and sellers since they usually do not have an objective viewpoint of the price distribution for the overall market of their interest. Drawing from data collected of all possible available properties for rent in Manhattan as of September 2023, this paper aims to strengthen our understanding of model residuals; specifically on machine learning models which generalize for a majority of the distribution of a well-proportioned dataset. Most models generally perceive deviations from predicted values as mere inaccuracies, however this paper proposes a different vantage point: when generalizing to at least 75\% of the data-set, the remaining deviations reveal significant insights. To harness these insights, we introduce the Price Anomaly Score (PAS), a metric capable of capturing boundaries between irregularly predicted prices. By combining relative pricing discrepancies with statistical significance, the Price Anomaly Score (PAS) offers a multifaceted view of rental valuations. This metric allows experts to identify overpriced or underpriced properties within a dataset by aggregating PAS values, then fine-tuning upper and lower boundaries to any threshold to set indicators of choice.
\end{abstract}

\section{Introduction}
\subsection{Rental Landscape of Manhattan}
The borough of Manhattan in New York City (NYC) has consistently ranked among the most expensive and competitive locations in the United States. With a population of 8.4 million in NYC, approximately 1.6 million individuals reside in Manhattan competing for 913,000 housing units available \cite{censusreporter, censusgov}. The competitiveness of the Manhattan housing market has been even more heightened due to by influx of individuals returning to the city post COVID, mainly due to the return to office policy of corporation \cite{nbcnews}. Simultaneously, the expiration of the 421-a tax abatement - a program that allows tax exemption for a period of time for condo and co-op - in June 2022 has led to a decline in the number of constructions, which has further worsened the housing supply \cite{thecitynyc}. Given the heightened competitiveness of the Manhattan rental market, there arises a crucial need for a reliable tool that can assist both tenants and landlords in identifying properties that are overpriced and underpriced. Such a tool would empower the the market participants to make informed decisions when looking for or putting up a place to rent, ensuring that opportunities are maximized.
 
\subsection{Value of Real-time Data in Real Estate Analysis}
Accurate and current market data plays a pivotal role in the precise estimation of rental prices. Real-time market data provides necessary granularity to incorporate the market dynamics into the model. By leveraging real-time information, predictive models can offer more valuable insight into the rental price estimation through adapting swiftly to evolving circumstances, thus enhancing their reliability and relevance. Especially for a volatile and dynamic market such as Manhattan, real-time data is even more beneficial in ensuring that the predictive models remain responsive to nuances of the market, fostering their efficacy as valuable tool to assist market participants to make informed decisions. 
  
\subsection{Main Objective}
This paper aims to introduce a new approach for identifying overpriced and underpriced rental properties by integrating real-time data scraped from StreetEasy and introducing a novel metric grounded in prediction residuals. In contrast to conventional approaches, our methodology embraces the dynamic nature of real-time data, ensuring a more responsive and up-to-date analysis. Furthermore, the research introduces a metric based on prediction residuals, enhancing the precision of identifying deviations from fair market values. Through identifying the machine learning models that could provide the most accurate prediction of the rental prices, the paper will also provide insight into the most important factors that influence the rental prices in Manhattan and explore variations in prices across different neighborhoods in the area. 

\section{Related Work}
In recent years, there have been a lot of academic interest in the use of big data and machine learning techniques in real estate price and rental prediction. Chen et al. (2016) discussed two primary advantages of machine learning over traditional statistical approaches \cite{chen2016}. Firstly, unlike statistical approaches, machine learning methods are more flexible and often do not require assumption on data distribution. Secondly, machine learning methods are able to capture the higher-order interactions and non-linear relationships in the data, and as a result, produce better predictions.  

Several studies have explored the use of machine learning in estimating rental prices. Zhou et al. (2019) tested several machine learning models (e.g., random forest, CART, bagged decision trees) and deep learning models (e.g., convolutional neural networks, recurrent neural networks) on the Atlanta rental data from Craiglist \cite{zhou2019}. The study found that combing textual information of the housing price, such as neighborhood, amenities with housing attributes would advance the understanding of the rental market dynamic. Zhu et al. analyzed a sample of approximately 49,000 Airbnb listings in NYC to identify the determinants of listing prices \cite{9253078}. The authors studied several machine learning methods, such as deep neural networks, generalized additive models, XGBoost, and bagging. The study concluded that deep neural networks with high complexity might cause over-fitting and impact model's robustness. Among all the methods, the bagging approach is strongly recommended as it combines the benefit of multiple models while minimizing the dominating power of any specific model. Dhillon et al. emphasized the importance of data cleaning and data pre-processing, followed by exploratory analysis to get a better understanding of the nature of the data prior to model building \cite{9376144}. The study then examined Linear Regression, Logistic Regression and Random Forest in predicting Airbnb prices in 28 most popular cities across the US. Random Forest was concluded to be the best model based on RMSE.  

Although numerous studies have addressed the prediction of rental prices, limited academic attention has been directed towards the  task of identifying overpriced and underpriced properties. The predominant focus in the existing researches lies in determining the optimal approach for rental price estimation, often neglecting the consequential step of utilizing the model outcomes to discern overpriced and underpriced properties. This gap in the literature underscores the need for a comprehensive exploration of not only accurate price estimation but also the practical application of model outcomes in real-world. Understanding the implications of property valuation beyond mere estimation is essential for informing strategic decision-making in the real estate and unlocking the full potential of predictive modeling

Traditionally, the conventional method for assessing whether the rental price of a property is above or below the fair market value involves comparing it to similar properties in the neighborhood. However, the diverse features associated with a real estate property can overwhelm prospective users, hindering their cognitive ability to analyze all the information and accurately gauge the actual market rental rate. During property research, individuals often rely on personal experiences, interests, and sometimes emotion to form their estimates. Despite the availability of numerous online platforms for property leasing, it remains challenging for users to find two comparable real estate properties for meaningful comparisons. Recognizing these limitations and the inherent difficulty users face, Mijatovi introduced the use of residuals between predicted and actual rental price to identify overestimated and underestimated properties. Under this approach, residuals are normalized by Z-score, which is further discretized in 5 groups indicating if the rental price is good, fair or high \cite{Mij2020}. To the best of our knowledge, this work represents a unique contribution, as we have not identified any other research that employs statistical metrics for the same purpose.

\section{Methodology}

\begin{figure}[h]
  \centering
  \includegraphics[width=\linewidth]{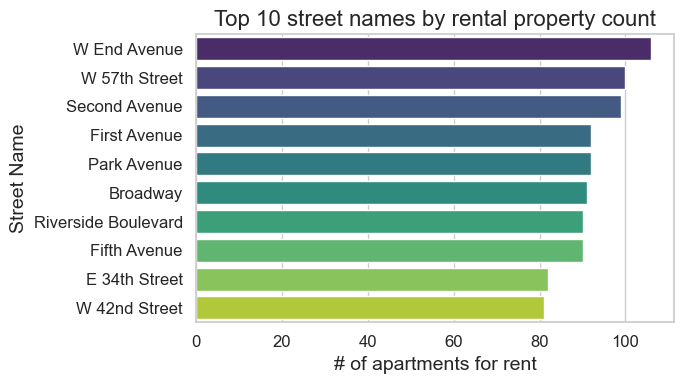}
  \caption{The top 10 streets with the most rentals listed}
  \label{fig:top10street}
\end{figure}

\subsection{Data Collection and Pre-Processing}
Historically, studies focusing on real estate markets have predominantly utilized national or local housing surveys such as American Community Survey data. However, these sources are often deficient in offering temporal and spatial specificity, which restricts their effectiveness in monitoring immediate fluctuations within local markets. With the migration of rental housing advertisements to online platforms, an increasing corpus of scholarly work has turned to web scraping methodologies to analyze rental market dynamics. Many of these studies have primarily relied on Craigslist as a source of data \cite{costa2021}. However, the streamlined listing process on Craigslist, characterized by minimal validation or verification procedures, results in a proliferation of fraudulent or deceptive postings. Therefore, this study will use data from StreetEasy, a prominent and popular online real estate listing website. The listings on StreetEasy are constantly monitored and verified for accuracy, which helps ensure that the information is up-to-date and factually correct \cite{streeteasy}. In addition, given the high data quality it resembles the actual population of Manhattan allowing for very valuable experimentation.

The StreetEasy website contains the majority of properties for rent in all boroughs of New York City, for this data collection process the focus was specifically on Manhattan. Based on a custom-built web-scraping script that uses selenium, the data of all 773 pages of apartments for rent in Manhattan, or 8,273 properties was collected. Initial features derived were title, address, price, details, and listing\_by. These features were then post-processed to derive additional features like location, rental\_type, beds and baths. Street names were also parsed out and used to see if features would be deemed useful in the model training process as shown in Figure~\ref{fig:top10street}.

\subsection{Primary Data Inspection}

\begin{figure}[htbp]
  \centering
  \includegraphics[width=\linewidth]{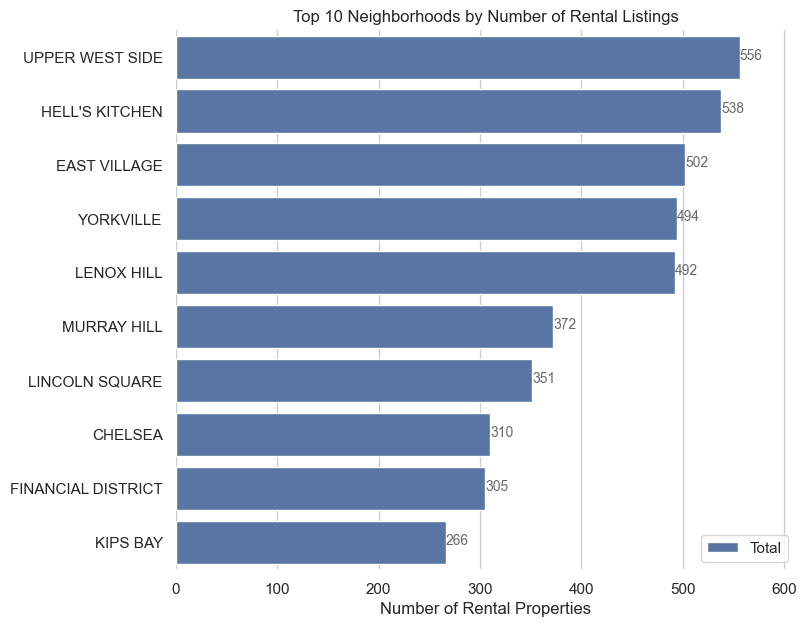}
  \caption{The top 10 neighborhoods based on the number of rental property listings in Manhattan. The Upper West Side has the highest number of available properties for rent as of September 2023.}
  \label{fig:top10neighborhoods}
\end{figure}

\begin{figure*}[htbp]
  \centering
  \includegraphics[width=\textwidth]{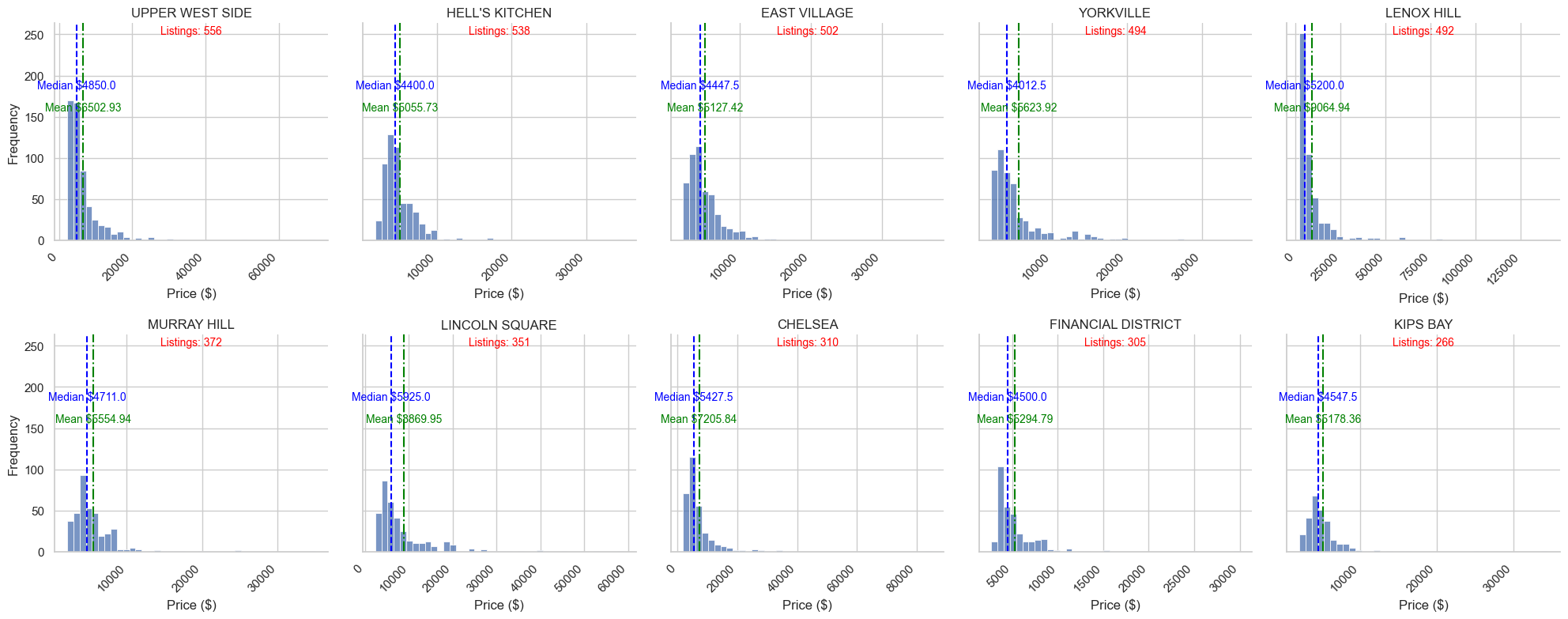}
  \caption{The top 10 neighborhoods' rental distributions. Median and mean prices vary showing skewed distributions.}
  \label{fig:top10dists}
\end{figure*}

As shown in Figure~\ref{fig:top10neighborhoods}, the Upper West Side stands out as the neighborhood with the largest number of listings, reflecting its popularity and density amongst rental properties. Figure~\ref{fig:top10dists} illustrates the distribution of rental prices across the top 10 neighborhoods out of all 53 that have the most listings. Out of the top 10 popular neighborhoods, the highest median rental price is in Lincoln Square. Lincoln Square also has the highest deviation between median and mean price; in addition to the right skewed distribution this suggests that there are a number of luxury rental properties that are significantly driving up the average rental price, indicating a market with a wide range of property values and likely some very high-end options.

When trying to understand the relative distributions between rental property bedroom size, it can be shown that there are significant outliers within the data in Table~\ref{fig:tableI} when looking at the maximum rental prices per bedroom size. A 7-bedroom rental has a maximum price of \$175,000 per month and minimum of \$34,000, while somewhere out there a 1-bedroom at its peak price in Manhattan is going for \$38,000 per month. When taking a deeper look, there are two condos at this maximum price for rent both of which residing on Fifth Ave, which makes sense given that Fifth Avenue is a prominent location in Manhattan. Overall, given the summary statistics it can be concluded that the dataset represents the relative market landscape well given the counts of the types of rentals and also accounts for those properties which deviate significantly from the majority.
\begin{table}[htbp]
\centering
\caption{Summary Statistics of Rentals by Price (\$)}
\label{tab:summary_stats}
\begin{tabular}{@{}lp{.5cm}p{.7cm}p{1cm}p{1cm}p{.8cm}r@{}}
\toprule
Bedrooms & Min & Max & Median & Mean & Std Dev & Count \\
\midrule
Studio & 1650 & 19500 & 3495.00 & 3463.61 & 1090.04 & 1360 \\
1 Bed & 1700 & 38000 & 4250.00 & 4395.46 & 1853.29 & 2804 \\
2 Beds & 1900 & 45000 & 5781.50 & 6782.34 & 4259.13 & 2448 \\
3 Beds & 2450 & 140000 & 7595.00 & 10678.33 & 9638.21 & 1177 \\
4 Beds & 2895 & 95000 & 9950.00 & 15522.33 & 14540.46 & 366 \\
5 Beds & 4300 & 150000 & 15622.50 & 25689.68 & 22960.37 & 90 \\
6 Beds & 5995 & 105000 & 37247.50 & 38793.00 & 28448.10 & 18 \\
7 Beds & 34000 & 175000 & 50000.00 & 69333.33 & 43072.61 & 9 \\
8 Beds & 53000 & 53000 & 53000.00 & 53000.00 & NaN & 1 \\
\bottomrule
\label{fig:tableI}

\end{tabular}
\end{table}
\subsection{Feature Engineering and Model Selection}
In preparation for modeling feature selection and engineering was performed to have adequate independent continuous and discrete variables to model the price which is a continuous dependent variable. Out of all of the features in the current dataset, unique parsed street names were not used as a feature, since there were 600 unique street names this could add a curse of dimensionality to the feature space. The features selected were beds, baths, location and rental\_type. Out of the selected features, beds and baths were converted to continuous features while location and rental type were one hot encoded as discrete ensuring to drop the first category of the variables to reduce correlations amongst variables for a total of 61 features.

As a starting point, linear regression was fitted to the dataset on a log transformed dependent variable and without to understand whether the data would follow the assumptions. For the purpose of making inferences on coefficient values and their statistical significance, it was found that the data did not fit these assumptions and thus linear regression was not selected. Instead of using linear models to derive feature impact, SHAP analysis was used to perform the same type of inference one could make similar to linear regression coefficient values. After testing many other models utilizing 10-fold cross-validation on 80\% of the dataset retaining a 20\% holdout set; gradient boosting trees (XGBoost) was selected. Gradient boosting trees combine many weak learned decision trees in a sequential manner to combine overall prominent predictions. XGBoost is an optimized version which includes additional regularization. It was used as the main model to base feature importance and analyze residuals for the purpose of identifying underpriced or overpriced properties based on the magnitude of the residuals using PAS values mentioned later. As shown in Table~\ref{tab:model_metrics}, the gradient boosting trees had the highest $R^2$ signifying that the independent variables explain approximately 79\% of the variance or variability represented in the rental prices overall.

\begin{table}[htbp]
\centering
\caption{Comparison of Machine Learning Model Metrics}
\label{tab:model_metrics}

\begin{tabular}{@{}lcccc@{}}

\toprule
Model & {MAE} & {MSE} & {$R^2$} & {RMSE} \\
\midrule
XGBoost & 0.206548 & 0.077212 & 0.793716 & 0.277469 \\
SVM & 0.204301 & 0.077224 & 0.793691 & 0.277487 \\
MLP & 0.209401 & 0.079272 & 0.789329 & 0.280020 \\
RandomForest & 0.210437 & 0.081370 & 0.782922 & 0.284679 \\
DecisionTree & 0.214435 & 0.087472 & 0.768681 & 0.293217 \\
GBM & 0.225018 & 0.087815 & 0.765316 & 0.295947 \\
Ridge & 0.223792 & 0.090514 & 0.758719 & 0.300196 \\
KNN & 0.225029 & 0.091859 & 0.754174 & 0.302781 \\
\bottomrule
\end{tabular}
\end{table}

\subsection{Feature Importance}

\begin{figure}[htbp]
  \centering
  \includegraphics[width=0.5\textwidth]{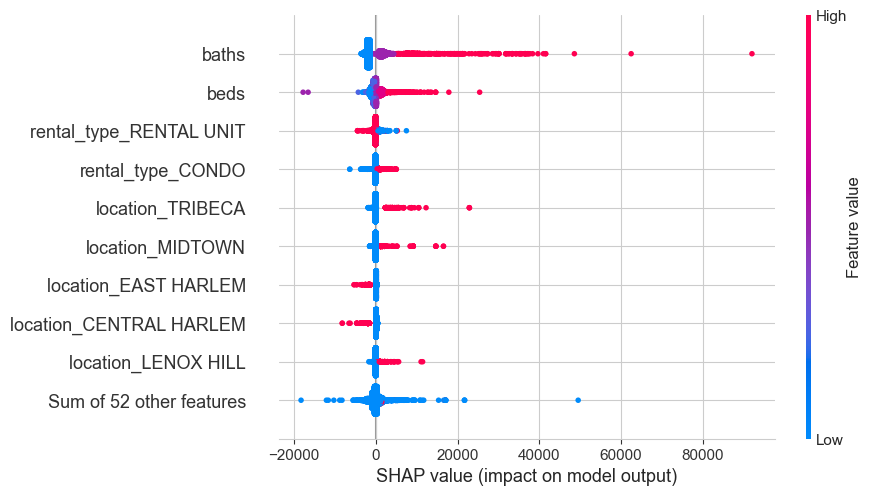}
  \caption{a SHAP analysis of the top features impact on price}
  \label{fig:top9shap}
\end{figure}

In order validate factors that may indicate a property of interest due to price discrepancies that are beyond the overall distribution, further confirmation among the feature impact on price and its alignment with the real world setting is conducted. An example to understand whether feature distributions resemble a well proportioned dataset relative to the market is validated by cross-checking with other literature. Upon analyzing the features which contribute to the price prediction in the selected model, it can be shown in Figure~\ref{fig:top9shap} as the value of baths or beds increases, there is a high increase on the SHAP value which is a direct impact on model output in this case price. The XGBoost model is suggesting that rentals in Tribeca or Midtown have a much higher impact on the price as opposed to East Harlem or Central Harlem. This suggests that rental units in East or Central Harlem are much less than those in Tribeca or Midtown in terms of the variance that the model learns when it comes to the overall distributions contributing to price. Although not mentioned in the above figure, South Harlem, Hamilton Heights, Yorkville, Hell's Kitchen and Washington Heights also have negative SHAP values, indicating lower price impact from rentals which reside in these locations. The observed outcome is consistent with the prevailing understanding of rental prices in Manhattan. As indicated by Corcoran, one of the prominent residential brokerages in New York City, Soho/Tribeca emerged as the highest-priced neighborhood for rental properties in October 2023 \cite{cocoran}. Concurrently, Washington Heights, Hamilton Heights, and East Harlem were identified as Manhattan neighborhoods ranking among the top 10 most affordable areas in New York City, according to information reported by Timeout \cite{timeout}.

\section{Identifying underpriced and overpriced properties using residuals}
\subsection{Capturing residuals of the full dataset}
Unlike most machine learning approaches that focus on the overall error of the test set to ensure that the model is performing well on what is unseen, here we focus on the overall error of all apartments. After hyperparameter tuning to find the most optimal model given the highest R-squared and lowest RMSE as mentioned earlier, the optimal model is then fit on the entire dataset without splitting the data. The catch here is to find the optimal model which predicts the unseen data utilizing cross-validation, then to use those optimal model parameters to refit to the entire dataset to then analyze the residuals holistically. Once this has been achieved, the residuals can be analyzed and the Price Anomaly Score can be applied for a full analysis of the dataset.

\subsection{Price Anomaly Score}
In this section, we introduce the Price Anomaly Score (PAS), designed to capture the extent of deviation from expected prices, indicating potential overvaluation or undervaluation of properties. This score utilizes the concept of Z-score application from previous literature to improve on identifying pricing discrepancies; it incorporates the magnitude, Z-score and fine tuning of boundaries from a parameter q.
\newline

\subsubsection*{Aggregation of PAS}
The Price Anomaly Score is mathematically defined for a set of properties with actual prices \(\mathbf{p}\) and predicted prices \(\mathbf{\hat{p}}\) obtained from a model. The PAS for each property \(i\) in the dataset, where \(i\) ranges from 1 to \(N\), is given by:

\begin{equation}
PAS_i = \left( \frac{p_i}{\hat{p}_i} \right) \times z_i
\end{equation}

Where:
\begin{itemize}
    \item \( \frac{p_i}{\hat{p}_i} \) is the relative price deviation for the \(i\)-th property, quantifying the magnitude of the pricing discrepancy in relative terms.
    \item \( z_i \) is the standardized residual (Z-score) for the \(i\)-th property, indicating how many standard deviations the residual of the \(i\)-th property is from the mean residual. It is calculated for each property \(i\) as:

    \begin{equation}
    z_i = \frac{Residual_i - \mu_{Residual}}{\sigma_{Residual}}
    \end{equation}

    Here, \(\mathbf{Residual}\) is the vector of residuals for all properties, \(\mu_{Residual}\) is the mean of these residuals, and \(\sigma_{Residual}\) is their standard deviation.
    \newline
\end{itemize}

\subsubsection*{Assumptions of PAS}
For the application of PAS, the following criteria should be met:
\begin{enumerate}
    \item The model should achieve a coefficient of determination \( R^2 \geq 0.75 \), indicating a substantial proportion of variance in the prices explained by the predictors.
    \item Implementation of cross-validation to ascertain the model's generalizability to unseen data and find the optimal hyperparameters.
    \item Evaluation of the model on a dataset partitioned into a training and testing split (80/20), with the test set containing no fewer than 1,000 samples. 
\newline
\end{enumerate}

\begin{figure*}[htbp]
  \centering
  \includegraphics[width=.95\textwidth]{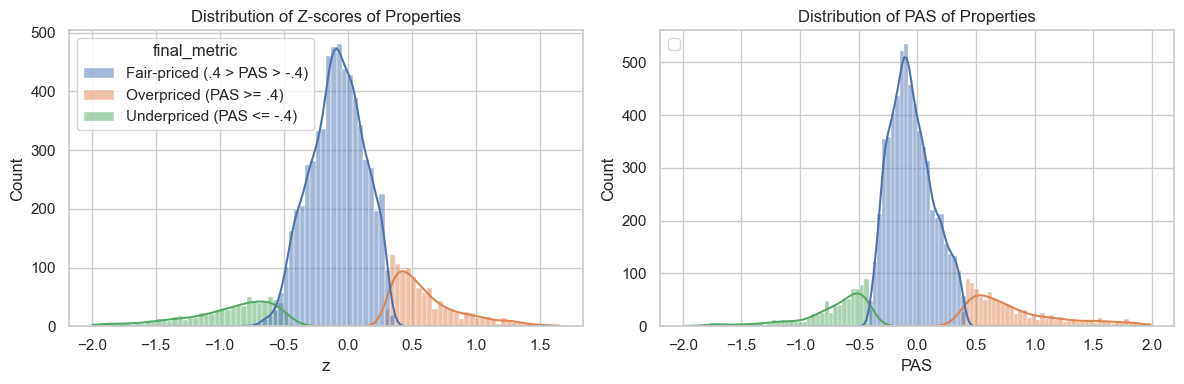}
  \caption{PAS shows a better fit to the normal distribution than the Z-score; q = 0.4}
  \label{fig:pasvsz2}
\end{figure*}

\subsubsection*{Price Category Classification}
Once the PAS is calculated for the entirety of the dataset following the criteria, pre-defined boundaries are set; these are adjustable according to the dataset's characteristics and application context. For the current study, we define the price categories as follows, with \( q \) representing the threshold parameter which fine-tunes the upper and lower bounds:

\[
\text{Price Category} = 
\begin{cases} 
\text{"Overpriced"} & \text{if } PAS \geq q \\
\text{"Underpriced"} & \text{if } PAS \leq -q \\
\text{"Fair-priced"} & \text{otherwise}
\end{cases}
\]

\begin{itemize}
    \item \( q \) is a scalar value which is selected by finding an example the user would identify as overpriced and identifying its PAS value, then associating a value which will fit all similar examples to it.
\end{itemize}

\subsection{Analysis of PAS versus Z-score}
In Figure~\ref{fig:pasvsz2}, the distribution of PAS values follows more of a normal distribution than the Z-score. When further comparing these two distributions, the valuation labeling is added in the legend to overlay the PAS boundaries using a q-value of 0.4 for both plots. This is conducted to show the counts of each metric relative to the PAS boundary selection. The tails of each distribution are significantly different which signals that a boundary based on a scalar value for Z-score alone may not capture all properties according to their market valuations. When using PAS, the importance of the error produced by the model is being scaled with an assumption that not all errors are created equal. If the Z-score is to be used alone to set boundaries for valuation, a small error for a very unusual property (high Z-score) could be considered more significant than the same error for a more typical property. This is due to the fact that the Z-score alone gives a measure of how far off a particular property's rental price is from the mean rental price, assuming a normal distribution of rental prices. However when utilizing the Price Anomaly Score, it signifies the predictive anomaly; how much the actual price differs from what the model predicts it should be, given all the factors the model takes into account. When PAS is used to set labels for valuation it scales the Z-score based on the error magnitude of the model, providing a different distinction between what is considered 'fair' priced and what is not. 

\begin{figure}[htbp]
  \centering
  \includegraphics[width=0.4\textwidth]{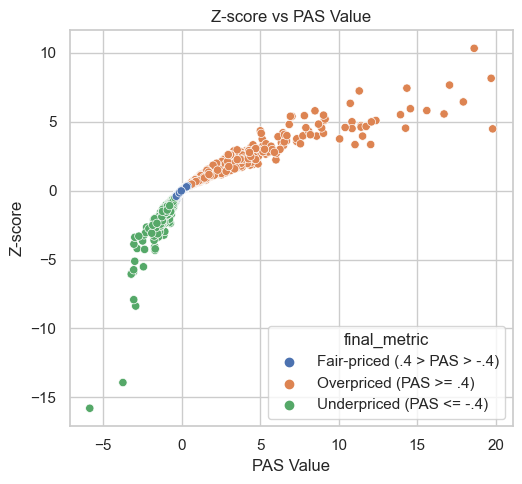}
  \caption{Each point represents a property in the dataset; q = 0.4}
  \label{fig:pasvsz}
\end{figure}

In Figure~\ref{fig:pasvsz} it can be shown that a logarithmic relationship exists between Z-score and PAS values. A larger Z-score magnifies the error making it much more significant when it comes to properties that are already outliers. Errors in prediction for properties that are closer to the average price (Z-score around 0) have less impact on PAS; this allows for more attention to be focused on outliers where model predictions are least accurate.

\section{Findings and Challenges}
\subsection{Implications of Findings}
For renters or real estate professionals, both the Z-score and PAS values have their positives and negatives for usage as metrics when identifying overpriced or underpriced properties. The Z-score is a purely statistical measure which is able to handle outliers in terms of price deviation from the mean, however it does not necessarily account for why a property might be an outlier. Using the Z-score for valuation labeling can be beneficial if the objective is to measure how far a property's residual deviates from the average of residuals in terms of standard deviations. However, when it comes to identifying the overall impact in the context of market dynamics (location, beds, baths) the Price Anomaly Score incorporates the predictive error which can suggest whether an outlier is statistically rare or genuinely mispriced in the housing market. 
\newline
\begin{figure}[htbp]
  \centering
  \includegraphics[width=0.45\textwidth]{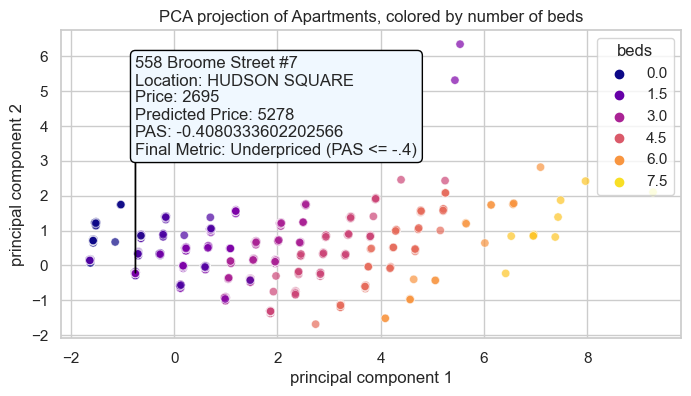}
  \caption{Identifying an underpriced 1-bedroom apartment}
  \label{fig:implications1}
\end{figure}

An example of PAS usage can be shown in Figure~\ref{fig:implications1} where the dimensionality of the dataset is reduced. Here it can be shown relative to the top two eigenvectors with the highest magnitude, the deviation from the predicted price can be visualized. This simplifies the dataset into a format where patterns can be more easily discerned. The properties are color-coded by the number of bedrooms; however the principal components do not suggest a strong clustering by this attribute alone. The highlighted property at "558 Broome Street \#7" serves as a case study showcasing a significant discrepancy between the market price and models prediction, resulting in a negative PAS value and its classification as 'Underpriced'. 

PAS values are heavily dependent on the relevance of the data fed into the predictive model. Both PAS and Z-score depend on the data however the PAS values effectiveness is tied to the performance and construction of the underlying model and quality of the data used in the model. The choice between these two metrics heavily depends on the specific requirements and constraints of the analysis or business objective. 
\subsection{Limitations and Challenges}
Despite the promising results obtained from the machine learning models used in this study, it is crucial to acknowledge several limitations and challenges that may impact the reliability and generalizability of the findings. Firstly, the study's scope is confined by the exclusive use of data from StreetEasy. This limits the analysis to the information available or provided by the platform and therefore does not include any real estate that is not rented on StreetEasy but may be transacted on another platform. Moreover, Manhattan's real estate markets, known for their dynamic characteristics, are susceptible to rapid changes influenced by economic conditions, external events, and other unpredictable factors, posing challenges for the selected model to adapt to sudden market shifts. Since the model uses data from a snapshot of data in time and therefore may not take into account some of these time-dependent factors. Additionally, Manhattan real estate may not be representative of real estate more broadly in the United States or the world, so the results of this study should not be generalized to broader areas.

Additionally, there are some challenges with the proposed residual feature metrics. The PAS metric relies on the assumption of normality in the distribution of the residuals. Therefore, if the rental prices do not follow a normal distribution, it may impact the validity of the metric and the subsequent classification of overvalued and undervalued properties. Lastly, the user-chosen threshold "q" for PAS introduces subjectivity into the determination of property valuation. The definition of what constitutes an over or underpriced real estate asset may be inherently subjective, and therefore the optimal threshold might vary based on user preferences or specific real estate market conditions. This study does its best to take this subjectivity into account by allowing for a moving threshold. In light of these constraints, it is essential to interpret the study results with these limitations in mind, recognizing the nuanced nature of real estate dynamics.

\section{Conclusion and Future Work}
The machine learning models employed in this study have shown promising results in predicting rental prices. The proposed PAS metric offers a novel approach to identifying overvalued and undervalued properties. However, the study is not without its limitations, as outlined in the previous section. Looking forward, the research could be enriched by incorporating  external factors such as neighborhood-specific  demographic and characteristics, proximity to amenities, and property condition, to complement the StreetEasy data. This addition could provide a more comprehensive understanding of the forces shaping rental prices and potentially contribute to the development of a more refined predictive model. Furthermore, the implementation of models capable of dynamic parameter updates in response to the evolving dynamics of the real estate market emerges as a promising avenue for future exploration. The integration of real-time data sources and adaptive learning algorithms could enhance the adaptability of models to sudden market shifts, ensuring a more accurate reflection of current condition. As future work explores the integration of external factors and development of dynamic models, it is crucial to emphasize that any models employed, alongside the user-defined threshold "q" should be subjected to continuous backtesting and cross-validation with the real world. This ensures the ongoing applicability and effectiveness of the models and their outcomes in the ever-evolving landscape of the real estate market.

\section*{Data Availability Statement}

The dataset supporting the conclusions of this article is available in the Zenodo repository. The dataset, titled \textit{StreetEasy Manhattan Properties September 2023}, was compiled by Youssef Sultan and is accessible at the following DOI: \href{https://doi.org/10.5281/zenodo.10207478}{10.5281/zenodo.10207478} \cite{sultan2023streeteasy}. This dataset encompasses a comprehensive collection of rental property listings in Manhattan, New York City, as of September 2023. The data was meticulously gathered through a custom-built web-scraping script utilizing selenium, targeting the StreetEasy website, known for its extensive and up-to-date listings across all boroughs of New York City. It contains post-processed and raw versions of the data collection.

{\small
\bibliographystyle{ieee_fullname}
\bibliography{egpaper_for_review}
}

\end{document}